\newif\ifabstract
\abstracttrue
\abstractfalse 
\newif\iffull
\ifabstract \fullfalse \else \fulltrue \fi

\documentclass[11pt]{article}
\usepackage{amsfonts}
\usepackage{amssymb}
\usepackage{amstext}
\usepackage[ruled,vlined]{algorithm2e}
\usepackage{amsmath}
\usepackage{xspace}
\usepackage{theorem}
\usepackage{graphicx}
\usepackage{url}
\usepackage{graphics}
\usepackage{colordvi}
\usepackage{colordvi}
\usepackage{subfigure}
\usepackage{lscape}
\usepackage{natbib}
\usepackage[english]{babel}
\usepackage{booktabs}
\usepackage{tabularx}
\usepackage{tikz}
\usetikzlibrary{positioning}

\advance \oddsidemargin by -0.8in
\newcommand{\myparskip}{3pt}
\parskip \myparskip

{\hspace*{\fill}$\Box$\par\vspace{4mm}}

\begin{document}

\title{Mobility Mode Detection Using WiFi Signals\footnote{\textbf{Published in the proceedings of IEEE International Smart Cities Conference 2018}}}
\author{Arash Kalatian\thanks{Laboratory of Innovations in Transportation (LiTrans), Ryerson University, Canada, Email: {arash.kalatian@ryerson.ca}} \and Bilal Farooq\thanks{Laboratory of Innovations in Transportation (LiTrans), Ryerson, Canada, Email: {bilal.farooq@ryerson.ca}}}

\begin{titlepage}
	\maketitle
	
	\thispagestyle{empty}
	
\begin{abstract}
We utilize Wi-Fi communications from smartphones to predict their mobility mode, i.e. walking, biking and driving. Wi-Fi sensors were deployed at four strategic locations in a closed loop on streets in downtown Toronto.  Deep neural network (Multilayer Perceptron) along with three decision tree-based classifiers (Decision Tree, Bagged Decision Tree and Random Forest) are developed. Results show that the best prediction accuracy is achieved by Multilayer Perceptron, with 86.52\% correct predictions of mobility modes.
\end{abstract}
	
\end{titlepage}

\section{Introduction}
Conventionally, customized surveys have been conducted to gather required sample. Intrinsic problems of traditional methods include involvement of human errors and biased responses, as well as being time-consuming, expensive and not representative. As a result, using location-aware technologies to collect data have gained popularity among scholars in different fields \citep{farooq2015ubiquitous}. These modern approaches in data collection have led to a dramatically growing interest in utilizing these data for mobility purposes.

Location aware technologies that have been implemented in mobility data collection processes include: body worn sensor, GPS data, GSM data, Wi-Fi data, Bluetooth transceivers data and RFID tags data. In general, these data collection approaches can be divided into two main categories: 1. User Centric Approaches and 2. Network Centric Approaches. User Centric Approaches, such as GPS or body worn sensors , require users to be actively involved in data collection procedure. Turning on device's GPS and installing and running a mobile application to collect GPS records for instance, make it hard to use these data in real life mobility problems. In Network Centric Approaches, such as GSM, Wi-Fi or Video Recording methods on the other hand, data can be collected passively with no intervention from users. 

Due to less infrastructure required, easier data preparation and calibration and more accuracy, GPS, GSM and Wi-Fi are the most referred source of data in related literature among other location-aware technologies: GPS-based approaches are the most precise sources of data in terms of geolocalization. The necessity of turning on mobile phones' GPS modules, however, makes these methods extremely battery draining. In addition, intervention from users is required (by turning on GPS modules or installing particular applications on their device), which prevents collecting data passively. Using cellular networks data resolves the problem of battery consumption and users intervention. Lack of accuracy in spotting user location, however, makes this source of data unreliable for estimating travel features in which accuracy plays an important role. By implementing sensors with the ability to collect connection information from  Wi-Fi enabled devices, location and movement data of users can be inferred passively with no intervention from users \citep{farooq2015ubiquitous}. Noted advantages of using Wi-Fi data led us to study the feasibility and accuracy of exploiting Wi-Fi data to detect mobility mode. 

Inferring mode of mobility is of interest as it gives insight into shares of different modes and their changes and trends over time. This information can be used in urban planning, contextual advertisement and designing bike lanes or sidewalks. Being able to extract mobility mode information in short period of time can help city decision makers to better prepare city infrastructure based on modal share. Other applications of inferring mobility mode include, but not limited to, carbon footprints of users with different modes, daily step count of users and estimation of traffic and pedestrian flow. 

Although studies have been conducted in this field, they have either focused on indoor environments, or have coupled Wi-Fi data with data from other sources. To the best of our knowledge, this is for the first time that mode detection is achieved using Wi-Fi as the only source of data on urban streets. Data used in this study are collected by URBANFlux system, consisting of OD\_Pod detectors. These detectors are able to record signals emitted by Wi-Fi enabled smartphones carried by most networks users. URBANFlux system is described in detail in \citep{farooq2015ubiquitous}. For the purpose of this research, eight detectors were installed on specific parts of downtown Toronto streets to collect Wi-Fi tracks of the participants walking, biking, or driving. Data collected in our experiment are used to train classifiers to predict mobility mode based on a set of defined features. Decision Tree, Bagged Decision Tree, Random Forest and Multilayer Perceptrons are used to predict participants mobility modes.

The rest of the paper is organized as follows: next section reviews previous works on the subject. We then describe our data collection and extraction procedure in detail. In the methodology section, features are explained and the algorithm implemented for selecting the most important ones is discussed. Classification techniques used in this study and their comparative analysis are then elaborated on. In the end, conclusions and future research plans are outlined.

\section{Background}
The general approach for determining mobility mode using smartphones in the existing literature consists of: (a) extracting related features from raw sensor data, (b) training an algorithm and (c) predicting mobility mode for the unseen data. 
GPS sensors have been the main source of data in related works, alone or along with additional sources. \cite{zheng2008understanding} use GPS data solely to detect mode of mobility. Using the approach suggested in this study, mobility mode of users in 76\% of the experiments are predicted correctly.

To add to the accuracy of the results, using multiple data sources has been widely practiced in related studies. \cite{reddy2008determining}, for instance, implement GPS data along with smartphones accelerometer data to distinguish users movements between walking, running, biking and motorized traveling . An accuracy of 93.6\% is reached using two-stage decision tree and discrete Hidden Markov Model. In another study, \cite{stenneth2011transportation}, add data from transportation network to determine users' mode of mobility between stationery, walking, biking, driving, and using public transit. Various methods including Decision Tree, Random Forest, Na\"ive Bayesian and Multilayer Perception are used in this study. Random Forest Method is chosen as the best classification method reaching an accuracy of 93.7\% .
\cite{dabiri2018inferring} utilized a convolutional neural network scheme to extract the mode of mobility using raw GPS data. They achieved an accuracy of 84\% in detecting walking, driving, biking, using bus and train.

In spite of the high accuracy in GPS-based methods, requiring users intervention, high battery consumption and difficulties for wide-spread implementations of such methods, have led scholars to use other, although less accurate, sources. The basic idea behind users' geolocalization with GSM data is to acquire their location based on the learning of which Based Transceiver Stations (BTS) they are connected to in specific time spans. \cite{sohn2006mobility}, for example, use coarse grained GSM data to determine whether a user is staying in a place, walking or driving. Boosted Logistic Regression is used in two phases and an accuracy of 85\% is reached in this study.  Low positioning accuracy, ping-pong handover effect and privacy concerns have been mentioned in this study as some of the main problems of using GSM data. In addition, relatively low density of BTS in some areas make GSM data an untrusted source for detecting mode specially at local levels. For trips within a block or neighborhood, for instance, GSM data cannot be used as a BTS cell size is at least 200 meters \citep{kalatian2016travel}.

Wi-Fi enabled devices can be discovered when they are inside Wireless Local Area Network. Being able to detect devices within access point's wireless range with no need of logging on, makes passive collection of data feasible. \cite{mun2008parsimonious} coupled Wi-Fi and GSM data to reach a classification accuracy of 87\% in urban areas. Features used for classification in this experiment include: Wi-Fi signal strength Variance, duration of dominant Wi-Fi access point, number of cell IDs that device connects to and residence time in cell footprint. A decision tree is trained for the purpose of this study.

Considering the advantages of exploiting Wi-Fi data to detect mobility mode, along with the fact that past studies with same source of data have mainly focused on limited indoor movements, or have added data obtained from other sources, make us explore the feasibility of using W-Fi data as the only source of data to detect mobility mode.

\section{Data}
Data used in this study is collected by URBANFlux system consisting of a network of detectors (named OD\_Pods) as described in \citep{farooq2015ubiquitous}. Every phone and computer has a unique hardware code, known as Media Access Control (MAC), by which it is identified. Wi-Fi enabled devices broadcast their MAC addresses periodically in order to explore available Wi-Fi Networks. URBANFlux detectors record these MAC addresses along with signal strengths and times of connection for individual devices carried by users who are in their coverage zone. Coverage zone of these detectors can be approximated as an sphere with a radius of 50 meters. Therefore, it is assumed in this study that connection details of mobile phones are recorded by an OD\_Pod when their direct distance to that OD\_Pod is less than 50 meters. 

\begin{figure}
	\centering
	\includegraphics[scale=0.3]{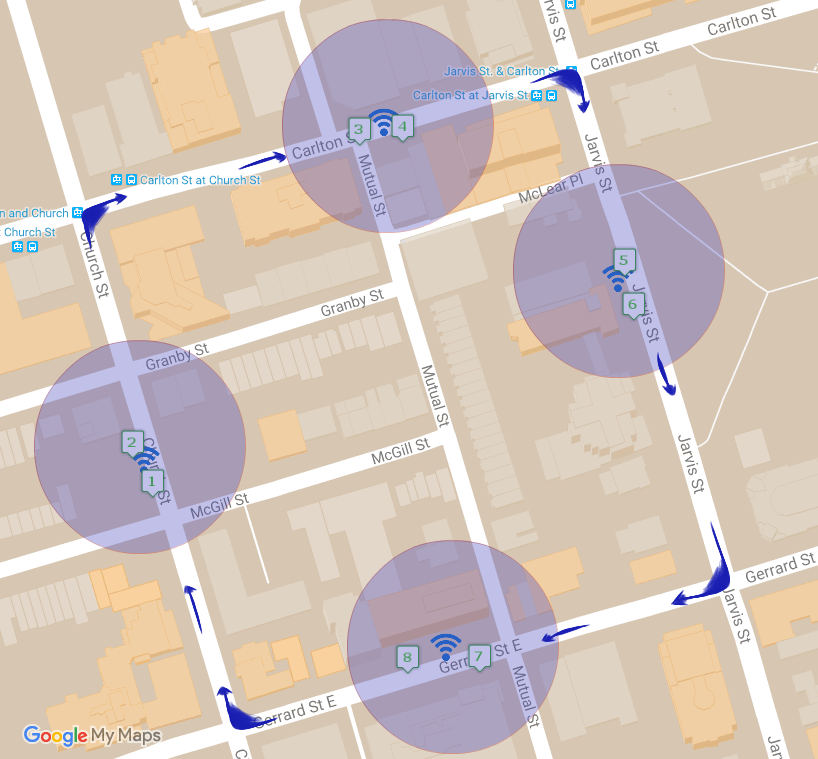}
	\caption{Experiment area and OD\_Pods location }
	\label{fig:odloc}
\end{figure}

Parts of four downtown Toronto streets forming a loop were selected for installing OD\_Pods. This area was selected in order for the experiment to be as realistic as possible. Data collection is conducted from 10 A.M. to 1 P.M on a weekday so that traffic on the streets resemble regular everyday life conditions. As it is depicted in Figure \ref{fig:odloc}, the selected parts form a grid loop with a perimeter of 857 meters. Coverage areas for OD\_Pods and direction of participants moving are also depicted in Figure\ref{fig:odloc}. Two OD\_Pods were installed in each location to decrease the chances of missing records due to possible OD\_Pods problems. To avoid signal overlaps, mid block points were chosen to install OD\_Pods. 

In this study, four lab members from LITrans were recruited for data collection procedure, each going around the designated loop for 10 rounds. Participants used different modes of mobility in a way that approximately equal number of data are recorded for each mode. Data recorded from movements between each two OD\_Pods are considered as a trip with their respective mode. As the data used in the analysis was related to recruited participants, ground truth modes of transport were known.     

By installing eight OD\_Pods at mid-block locations, travel times can be measured for participants. Data are collected from participants with different modes of mobility: walking, biking and driving. Having connection time and distance between OD\_Pods, a relative travel speed for participants is calculated as the normalized ratio of distance and time. Relative speed along with features such as average and variance of signal strengths and other derived features  are then used to infer mode of mobility of participants. The details of feature extraction and classification are further elaborated in next section.

\begin{table}
	\centering
	\caption{Total number of trips collected for each mode}
	\begin{tabular}{l c}
		\hline
		{\bf Mode} & {\bf Number of Trips} \\
		\hline
		\hline
		{Walking} & 142  \\
		{Biking} &        108     \\
		{Driving} &        150     \\
		\hline
	\end{tabular} 
	\label{tab:Tripsum}
\end{table}

\section{Methodology}
Intuitively, mobility patterns for different mobility modes are different. Users' modes can be classified based on these differences in the patterns.
In regard to algorithm performance, previous studies have suggested that the Decision Tree-based approaches perform well in terms of classifying these patterns. In addition, Deep Neural Networks (DNN) have been widely used and have performed successfully in various fields in recent years. In this section, we first discuss the importance of feature selection and variable importance ranking in classification. We then describe ``ReliefF algorithm'' 
implemented in this study for feature selection. Next, decision tree-based algorithms and DNN algorithm used for predicting mobility mode are discussed in detail. 
\subsection{ReliefF algorithm}
A feature selection algorithm should be utilized before using dependent variables 
as inputs of any machine learning algorithm for classification. The main reason for 
determining variable importance before training a classifier is making the 
calibration process of base learners more efficient. 
In this study, ReliefF algorithm is used to rank variables based on their 
importance. ReliefF algorithm was proposed by \cite{kononenko1997overcoming}
for multinomial classification.The basic idea of 
Relief Algorithms family is to estimate the quality of variables based on their 
ability to distinguish between observations that are close to each other. 

\subsection{Decision Tree}
Decision Tree (DT) classification uses a decision tree structure to predict class 
of observations based on their features. Basically, a tree is learned by testing 
attribute values, and splitting the source set to subsets based on that. For every 
derived subset, this procedure is repeated until no improvement is observed on 
predictions. The core algorithm for building decision trees employs a top-down, 
greedy search through the space of possible branches. Details of DT algorithm for 
classification can be found in the work of \cite{breiman1984classification}.
\subsection{Bagged Decision Tree}
To improve prediction accuracy of DT, ensemble learning algorithms are practiced. Fundamentally, an ensemble learning algorithm is using combined results of a set of classifiers. In this study, Bagged Decision Tree (BDT) is used to reduce the error of prediction for individual decision trees. In BDT, a number of Decision Trees are grown with samples from training data set. Samples are drawn from training set randomly and with replacement. Number of Decision Trees ($n$) should be optimally tuned. Error is reduced with increasing the number of trees , and remains stationary after an optimal value of $n$.

Algorithm \ref{Bag} describes the bagging algorithm used in BDT. Sample sizes are equal to the size of training dataset, so in bagging training set, some observations may appear several times. Each tree is calibrated individually and the predictions are combined by considering majority vote of trees. The idea behind growing multiple trees is that diversity among DT classifiers, compensates for the increase in error rate of each individual classifier.  
\IncMargin{1em}
\begin{algorithm}
	\SetKwData{Left}{left}\SetKwData{This}{this}\SetKwData{Up}{up}
	\SetKwFunction{Union}{Union}\SetKwFunction{FindCompress}{FindCompress}
	\SetKwInOut{Input}{input}\SetKwInOut{Output}{output}
	Set $D=\emptyset$ \\
	Set number of trees to be trained $n$\\
	\For{$i\leftarrow 1$ \KwTo $n$}{
		\emph{select a random sample $R_s$ with replacement from training set }\;
		\emph{train a Decision Tree $D_R$ based on $R_s$  }\;
		\emph{$D=D\cup D_R$}\;
	}
	Return D
	\caption{Bagging algorithm}\label{Bag}
\end{algorithm}\DecMargin{1em}

\subsection{Random Forest}
Another ensemble Decision Tree-based algorithm is Random Forest (RF), developed by \cite{breiman2001random}. In RF, in addition to bootstrap sampling of observations explained in BDT section, a subset of input variables are randomly selected as features for each tree. The pseudopod for RF algorithm is presented in Algorithm \ref{RF} \citep{liaw2002classification}. 
\IncMargin{1em}
\begin{algorithm}
	\SetKwData{Left}{left}\SetKwData{This}{this}\SetKwData{Up}{up}
	\SetKwFunction{Union}{Union}\SetKwFunction{FindCompress}{FindCompress}
	\SetKwInOut{Input}{input}\SetKwInOut{Output}{output}
	Set $D=\emptyset$ \\
	Set number of trees to be trained $n_t$\\
	Set number of features to be used for each tree $n_f$ \\
	\For{$i\leftarrow 1$ \KwTo $n_t$}{
		\emph{select a random sample $R_s$ with replacement from training set }\;
		\emph{select $n_f$ random variables from all predictors: $R_f$}\;
		\emph{train a Decision Tree $D_R$ based on $R_s$ with features $R_f$ }\;
		\emph{$D=D\cup D_R$}\;
	}
	Return D
	\caption{Random Forest algorithm}\label{RF}
\end{algorithm}\DecMargin{1em}
To predict the class of new data, predictions from $n_t$ trees are aggregated by considering majority vote. 
As it is shown in Algorithm \ref{RF}, two parameters should be optimally tuned: number of trees($n_t$)  and number of features ($n_f$).
\subsection{Multi-layer Perceptron}
Introduced in \cite{rumelhart1986general}, Multilayer Perceptron (MLP) is a feed-forward artificial neural network. MLP consists of an input layer on one end, an output layer on the other end and an arbitrary number of hidden layers in between. The information start in the input layer and propagate through the network layer by layer (\cite{honkela2001nonlinear}). In the mathematical form, each Perceptron can be written as in equation \ref{eq}. 
\begin{equation}\label{eq}
y=\varphi * (W^T*x+b)
\end{equation}
where $\varphi$ is the activation function, W is the vector of weights, b is the bias and x is the vector of inputs.

The perceptron forms a linear combination of inputs and their weights to compute an output, and puts the output in a nonlinear activation function. The perceptrons will then be used as parts of a more complex structure in MLP. In the MLP architecture, number of nodes in input layer and output layer represent number of features and number of labels (modes in our study) respectively.

MLPs have been used successfully in different areas, such as  speech recognition, image recognition, and machine translation. For the purpose of this study, we utilize a MLP scheme for mode detection using Keras Package in R \citep{chollet2017kerasR}. 
\section{Results}
\subsection{Features}
Having MAC addresses of participants' devices, connection data belonging to each participant can be separated. Every movement between two OD\_Pods is considered as a trip observation, which includes data from origin OD\_Pod and destination OD\_Pod. Based on connection details from origin and destination OD\_Pods, 15 variables are extracted as possible features for classification. These variables can be categorized in 4 groups in general, which are: 
\begin{enumerate}
	\item Time Related Variables: Relative Travel Speed and connection time variables are included in this group. Relative Travel Speed is represented by the normalized ratio of ``distance with no coverage zone between two involved OD\_Pods'', to its respective travel time. Speed related variables have been the main feature for classification in related literatures. However, using speed solely does not guarantee satisfactory results. This can be explained by the fact that in congested areas, different mobility modes move with similar speeds. 
	
	\item Connection time variables represent variables related to the time that a device is discovered in the coverage zone of an OD\_Pod.
	\item Connection Number Variables: number of connections to an OD\_Pod while a device is in its coverage zone. Intuitively, when a user spends more time in a coverage zone, for example walks in a coverage zone, number of connections increases.
	\item Signal Strength Variables: Variance, first and second derivative of signal strengths during the time a device is connected to an OD\_Pod.
\end{enumerate}
\subsection{Variable Importance}
Using ReliefF algorithm, a ranking of variables based on their weighted importance is obtained. Variables with their respective units, ranks and weight importances are presented In Table \ref{tab:varrank}. As it can be inferred from the table, variables regarding number of connections are the most relevant features for determining mobility mode. When participants move with a slower speed, their device spends more time in a coverage zone, resulting in more number of connections and time. As expected, relative speed is another key variable for training a classifier. As it is presented in Table \ref{tab:varrank}, all the weights calculated have positive values, which means all of them should be taken into account for classification.     
\begin{table}
	\centering
	\caption{Details of features defined for classification and their weight importance and ranking}
	\begin{tabular}{|l|l|l|l|l|}
		\hline
		{\bf ID} & {\bf Variable}& {\bf Unit} & {\bf Rank}&{\bf Weight} \\
		\hline\multicolumn{ 5}{|l|}{\it Time } 
		\\
		\hline
		{1} & Relative Travel Speed  & - & 4 & 0.050  \\
		\hline
		{2} &        Origin Connection Time  & s & 7 &0.033 \\
		\hline
		{3} &        Destination Connection Time & s & 9&0.020   \\
		\hline\multicolumn{5}{|l|}{\it Connection Number}\\
		\hline
		{4} &        Number of Connections-Orig & - &2&0.073   \\
		\hline
		{5} &        Number of Connections-Dest & - &3 &0.059  \\
		\hline
		{6} &        Number of Connections-Avg  & - &1&0.104 \\
		\hline\multicolumn{5}{|l|}{\it Signal Strength}\\
		\hline
		{7} &        Signal Strength Variance-Orig & \(\displaystyle dBm^2\) &11&0.010 \\
		\hline
		{8} &        Signal Strength Variance-Dest & \(\displaystyle dBm^2\)&10&0.010   \\
		\hline
		{9} &        Signal Strength Variance-Avg & \(\displaystyle dBm^2\)&6&0.036   \\
		\hline
		{10} &        Signal Strength 1st Derivative-Orig & dBm/s &5&0.036    \\
		\hline
		{11} &        Signal Strength 1st Derivative-Dest & dBm/s&14&0.003     \\\hline
		{12} &        Signal Strength 1st Derivative-Avg & dBm/s&8&0.021   \\\hline
		{13} &        Signal Strength 2nd Derivative-Orig& \(\displaystyle dBm/s^2\)&13&0.006   \\\hline
		{14} &        Signal Strength 2nd Derivative-Dest&\(\displaystyle dBm/s^2\) &15&0.0002  \\\hline
		{15} &        Signal Strength 2nd Derivative-Avg & \(\displaystyle dBm/s^2\)&12& 0.0009    \\\hline
	\end{tabular} 
	\label{tab:varrank}
\end{table}
\subsection{Classification}
Classification process starts with training a DT. In spite of easy calibration process, DT does not result in satisfying accuracy. The minimum number of leaf node observations and branch node observations are set to their default value, 1 and 10 respectively. 
Confusion matrix for test data for DT is presented in Table \ref{tab:ConMatDT}. In this table, actual number of trips is compared to DT predictions for different modes of travel. Column \textbf{Recall} for each mode indicates the number of correct predictions for that mode divided by total number of trips with that mode. In row \textbf{Precision}, number of correct predictions for each mode is divided by total number of times that a mode is predicted. As it is provided in Table\ref{tab:ConMatDT}, biking is the least accurately predicted mode with a recall of 47.91\% and precision of 60.52\%. Total number of correct predictions in DT appears to be 68.75\%. 
\begin{table}
	\centering
	\caption{Confusion Matrix for Decision Tree}
	\begin{tikzpicture}[
	box/.style={draw,rectangle,minimum size=1.5cm,text width=1.5cm,align=center}]
	\matrix (conmat) [row sep=.1cm,column sep=.1cm] {
		\node (1) [box,
		label=left:\bf Walking,
		label=above:\bf Walking,
		] {53};
		&
		\node (2) [box,
		label=above:\textbf{Biking},
		] {3};
		&
		\node (3) [box,
		label=above:\textbf{Driving},
		] {1};
		&
		\node(4) [label=above:\textbf{Total}]{57};
		&
		\node(5) [label=above:\textbf{Recall\%}]{92.98};
		\\
		\node (6) [box,
		label=left:\bf Biking,
		] {12};
		&
		\node (7) [box,
		] {23};
		&
		\node (8) [box,
		] {13};
		&
		\node(9) {48};
		&
		\node(10) {47.91};
		\\
		\node (11) [box,
		label=left:\bf Driving,
		] {9};
		&
		\node (12) [box,
		] {12};
		&
		\node (13) [box,
		] {34};
		&
		\node(14) {55};
		&
		\node(15) {61.81};
		\\
		\node (16) [
		label=left:\textbf{Total},
		] {74};
		&
		\node (17) {38};
		&
		\node (18)  {48};
		&
		\node (9)  {160};
		\\
		\node (10) [
		label=left:\textbf{Precision},
		] {71.62};
		&
		\node (8) {60.52};
		&
		\node (9)  {70.83};
		\\
	};
	\node [rotate=90,left=.1cm of conmat,anchor=center,text width=1.5cm,align=center] {\textbf{Actual}};
	\node [above=.1cm of conmat,align=center,anchor=center] {\textbf{Prediction}};
	\end{tikzpicture}
	\label{tab:ConMatDT}
\end{table}

To reduce the errors of DT and improve the classification prediction accuracy, BDT algorithm is implemented. To use BDT for classification, number of trees included in BDT, \textit{n}, should be calibrated. Different numbers of trees are tested in this study to calculate the optimal value of \textit{n}. The value of \textit{n} is set to be 10, 20, 50, 100, 200, 500, 1000, and it turned out that the prediction accuracy remains approximately constant for \textit{n} grater than 200. Confusion matrix of BDT for test data is provided in Table \ref{tab:ConMatBDT}. As the recall and precision values of this table suggest, implementing bagging strategy improves recall and precision for all modes, except a decrease in recall percentage of walking. In addition, the total prediction accuracy increases from 68.75\% for DT to 78.25\% in BDT. 
\begin{table}
	\centering
	\caption{Confusion Matrix for Bagged Decision Tree}
	\begin{tikzpicture}[
	box/.style={draw,rectangle,minimum size=1.5cm,text width=1.5cm,align=center}]
	\matrix (conmat) [row sep=.1cm,column sep=.1cm] {
		\node (1) [box,
		label=left:\bf Walking,
		label=above:\bf Walking,
		] {50};
		&
		\node (2) [box,
		label=above:\textbf{Biking},
		] {5};
		&
		\node (3) [box,
		label=above:\textbf{Driving},
		] {2};
		&
		\node(4) [label=above:\textbf{Total}]{57};
		&
		\node(5) [label=above:\textbf{Recall\%}]{87.72};
		\\
		\node (6) [box,
		label=left:\bf Biking,
		] {4};
		&
		\node (7) [box,
		] {36};
		&
		\node (8) [box,
		] {8};
		&
		\node(9) {48};
		&
		\node(10) {75.00};
		\\
		\node (11) [box,
		label=left:\bf Driving,
		] {9};
		&
		\node (12) [box,
		] {7};
		&
		\node (13) [box,
		] {39};
		&
		\node(14) {55};
		&
		\node(15) {70.91};
		\\
		\node (16) [
		label=left:\textbf{Total},
		] {63};
		&
		\node (17) {48};
		&
		\node (18)  {49};
		&
		\node (9)  {160};
		\\
		\node (10) [
		label=left:\textbf{Precision},
		] {79.37};
		&
		\node (8) {75.00};
		&
		\node (9)  {79.59};
		\\
	};
	\node [rotate=90,left=.1cm of conmat,anchor=center,text width=1.5cm,align=center] {\textbf{Actual}};
	\node [above=.1cm of conmat,align=center,anchor=center] {\textbf{Prediction}};
	\end{tikzpicture}
	\label{tab:ConMatBDT}
\end{table}

\begin{table}[!h]
	\centering
	\caption{Confusion Matrix for Random Forest}
	\begin{tikzpicture}[
	box/.style={draw,rectangle,minimum size=1.5cm,text width=1.5cm,align=center}]
	\matrix (conmat) [row sep=.1cm,column sep=.1cm] {
		\node (1) [box,
		label=left:\bf Walking,
		label=above:\bf Walking,
		] {50};
		&
		\node (2) [box,
		label=above:\textbf{Biking},
		] {4};
		&
		\node (3) [box,
		label=above:\textbf{Driving},
		] {3};
		&
		\node(4) [label=above:\textbf{Total}]{57};
		&
		\node(5) [label=above:\textbf{Recall\%}]{87.72};
		\\
		\node (6) [box,
		label=left:\bf Biking,
		] {3};
		&
		\node (7) [box,
		] {37};
		&
		\node (8) [box,
		] {8};
		&
		\node(9) {48};
		&
		\node(10) {77.08};
		\\
		\node (11) [box,
		label=left:\bf Driving,
		] {7};
		&
		\node (12) [box,
		] {2};
		&
		\node (13) [box,
		] {46};
		&
		\node(14) {55};
		&
		\node(15) {83.64};
		\\
		\node (16) [
		label=left:\textbf{Total},
		] {60};
		&
		\node (17) {43};
		&
		\node (18)  {57};
		&
		\node (9)  {160};
		\\
		\node (10) [
		label=left:\textbf{Precision},
		] {83.33};
		&
		\node (8) {86.05};
		&
		\node (9)  {80.70};
		\\
	};
	\node [rotate=90,left=.1cm of conmat,anchor=center,text width=1.5cm,align=center] {\textbf{Actual}};
	\node [above=.1cm of conmat,align=center,anchor=center] {\textbf{Prediction}};
	\end{tikzpicture}
	\label{tab:ConMatRF}
\end{table}

As the last DT based algorithm, mobility mode is predicted by training a RF algorithm. For RF, the recommended value of $n_f$ is one third of total number of predictors \citep{saadi2017investigation}, which means setting $n_f$=5. To ensure about the optimum values of $n_t$ and $n_f$, multiple numbers are set and tested for both parameters. Predictions of RF trained with different parameter values appears to be most accurate with $n_t$=400 and $n_f$=5, which confirms the recommendation of setting number of randomly selected predictors equal to one third of total number of features. The Confusion Matrix of RF for test data is presented in Table \ref{tab:ConMatRF}. A total prediction accuracy of 83.13\% is reached using RF algorithm. As it is presented in Table \ref{tab:ConMatRF}, recall and precision values for all modes is also improved, compared to results from BDT. Similar to previous methods, walking has the best recall percentage: 50 observations out of total 57 walking observations are predicted correctly. On the other hand, 10 biking or driving labeled observations are predicted as walking by mistake, which makes a precision of 83.33\%. This precision is still more precise than the results of DT and BDT. Both recall and precision for driving labeled observations are above 80\%, and greater than their respective values in two previous methods. 

\begin{figure}
	\centering
	\includegraphics[scale=0.28]{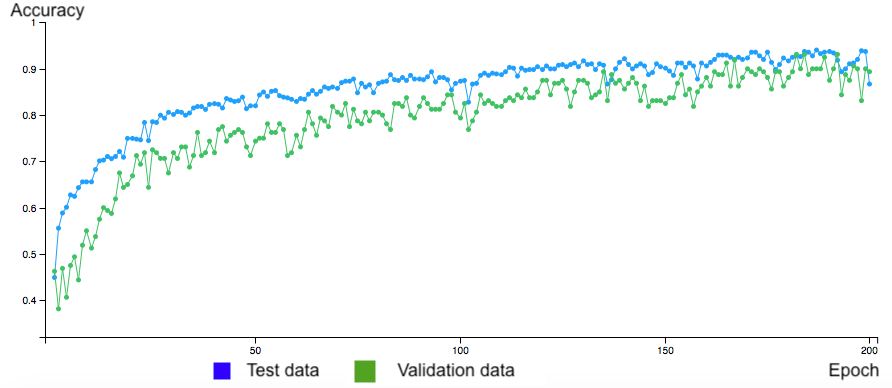}
	\caption{Accuracy of Training and Validation data on different epochs }
	\label{fig:acc}
\end{figure}

In the end, Multilayer Perceptron is used to predict mode of mobility. With multiple parameters to be tuned, it is more difficult to find an optimum MLP algorithm compared to DT based algorithm. These parameters include: Number of hidden layers, Number of nodes in each hidden layer, Optimization method, batch size and number of epochs. After implementing different architectures for the network, 87 percent accuracy in total predictions is achieved. Table \ref{tab:par} presents a summary of the final model attributes. As presented in this table, the number of epochs is set to be 200. An epoch is one complete presentation of the dataset to the learning algorithm. This number is set based on the accuracy of model is predicting validation and training datasets. The accuracy of the predictions for different number of epochs is presented in Figure \ref{fig:acc}. As it can be seen in this figure, No observable change occurs after 200 epochs. 
\begin{table}
	\centering
	\caption{Values for MLP parameters}
	\begin{tabular}{|l|l|}
		\hline
		{\bf Name} & {\bf Value} \\
		\hline
		{Number of Epochs} & 200 \\
		\hline
		{Optimization Method} &        ADAM   \\
		\hline
		{Number of hidden layers} &        4   \\
		\hline
		{Input and hidden layers} &        activation: RELU\\
		\hline
		{Number of nodes} &       all hidden layers: 15   \\
		\hline
		{Output layer} &      activation: softmax  \\
		\hline
		{Batch size} &      20 \\
		\hline
	\end{tabular} 
	\label{tab:par}
\end{table}
The confusion matrix for MLP algorithm is provided in Table \ref{tab:ConMatDL}. As it can be inferred from the table, comparing to all previous DT based algorithms, Recall and Precision have increased significantly. Driving remains the least accurate mode, with around 15 percent of driving mode predicted as other modes. This error could mainly be a result of congested traffic conditions at the intersections, which may make it hard to distinguish between different modes. Comparing to Random Forest algorithm, the total accuracy improved by 3 percent, which may be further be improved by using other deep learning algorithms.
Finally, biking has the least recall percentage, mainly due to large number of biking observations predicted as driving. This error probably reflects the fact that in a congested area such that the data collection of this study was conducted, bikes and cars may share speed related features. 
\begin{table}
	\centering
	\caption{Confusion Matrix for Multilayer Perceptrons}
	\begin{tikzpicture}[
	box/.style={draw,rectangle,minimum size=1.5cm,text width=1.5cm,align=center}]
	\matrix (conmat) [row sep=.1cm,column sep=.1cm] {
		\node (1) [box,
		label=left:\bf Walking,
		label=above:\bf Walking,
		] {52};
		&
		\node (2) [box,
		label=above:\textbf{Biking},
		] {1};
		&
		\node (3) [box,
		label=above:\textbf{Driving},
		] {4};
		&
		\node(4) [label=above:\textbf{Total}]{57};
		&
		\node(5) [label=above:\textbf{Recall\%}]{89.47};
		\\
		\node (6) [box,
		label=left:\bf Biking,
		] {1};
		&
		\node (7) [box,
		] {40};
		&
		\node (8) [box,
		] {7};
		&
		\node(9) {48};
		&
		\node(10) {83.33};
		\\
		\node (11) [box,
		label=left:\bf Driving,
		] {3};
		&
		\node (12) [box,
		] {5};
		&
		\node (13) [box,
		] {47};
		&
		\node(14) {55};
		&
		\node(15) {85.45};
		\\
		\node (16) [
		label=left:\textbf{Total},
		] {56};
		&
		\node (17) {46};
		&
		\node (18)  {58};
		&
		\node (9)  {160};
		\\
		\node (10) [
		label=left:\textbf{Precision},
		] {92.86};
		&
		\node (8) {86.96};
		&
		\node (9)  {81.03};
		\\
	};
	\node [rotate=90,left=.1cm of conmat,anchor=center,text width=1.5cm,align=center] {\textbf{Actual}};
	\node [above=.1cm of conmat,align=center,anchor=center] {\textbf{Prediction}};
	\end{tikzpicture}
	\label{tab:ConMatDL}
\end{table}

\section{Conclusions }
In this paper, we used Wi-Fi communication data for determining individual's mobility mode. URBANFlux technology \citep{farooq2015ubiquitous} is used in this study to record MAC addresses along with signal strengths and time-stamp of communication from Wi-Fi enabled smartphones. Eight detectors were installed in four different locations in a designated loop route in downtown Toronto and four participants walked, biked and drove this route. Each movement between two OD\_Pods in different locations is considered as an observation in the dataset, leading to a total of 400 observations.

Fifteen features are derived based on times and speeds, signal strengths and number of connections of observations. ReliefF algorithm is used to estimate the quality of predictors for feature selection. Three decision tree-based algorithms and one Deep Neural Network algorithm classifiers are trained and travel modes for test data are predicted. Calibrating and training a Multilayer Perceptron algorithm resulted in an accuracy of 86.52\% in mode classification: 89.47\% for walking, 83.33\% for biking and 85.45\% for driving mode.

Unlike other related works on this topic, in this study Wi-Fi data are not coupled with other sources of data to detect mobility mode in urban areas. This makes our proposed approach more cost-effective and easier to implement--with no interventions from users required. As long as smartphone Wi-Fi is on, which is almost always the case, we are able to detect users carrying them in the proximity of our detectors. Another key differentiating feature in this study is the implementation of Wi-Fi based mode detection on actual urban roads with real traffic at reasonably large scale. Regarding the accuracy in prediction, our study performs well comparing to other studies using solely Wi-Fi data. The highest accuracy in related studies can be obtained by 
This method can also be used within URBANFlux system by city decision makers and planners to have a better understanding of users travel habits and their trends over time. City of Toronto, for instance, is collecting and publishing travel times, counts and travel speeds data of vehicles from Bluetooth and WiFi sensors on streets and highways across the city (\citeyear{opendata}). The same infrastructure can be simply improved to incorporate mode detection algorithms for estimation of modal in different parts of the city, with no additional burden on the city's economy. This method can also be used in a real-time within URBANFlux system to acquire information like traffic conditions or level of service for bike lanes and sidewalks.  

In the context of future work, this study can be improved in different aspects. In terms of methodology, other feature selection algorithms and classification methods can be explored to compare the results with the current study. Taking heterogeneity of smartphones into account may also add to the precision of this study. Other travel modes, such as buses, subways or stationary mode can be investigated in future work. Considering the large unlabeled dataset that has been collected during this study, the models can further be improved using semi-supervised learning algorithms.  In addition, dataset can be expanded to consist different parts of city, with more participants recruited to have a bigger and more comprehensive dataset. URBANFlux technology allows affordable longterm monitoring of network users \citep{alexandra2017urbanflux}, which provides perspective on exploring other traffic characteristics such as routes, origin-destination tables, activity patterns, and fundamental relationships between speed, density, and flow of traffic.

\bibliographystyle{plainnat}
\bibliography{temp_cites}

\end{document}